\documentclass[letterpaper, 10pt, conference]{ieeeconf}
\IEEEoverridecommandlockouts 
\usepackage[utf8]{inputenc}
\usepackage{hyperref}
\usepackage{amsmath}
\usepackage{amssymb}
\usepackage{graphicx}
\usepackage{dsfont}
\usepackage{todonotes}
\usepackage[T1]{fontenc}
\usepackage{textcomp}
\usepackage{subfig}
\usepackage{algpseudocode}
\usepackage{algorithm}
\usepackage{makecell}

\newcommand{\etal}{\textit{et al}.~}

\title{\bf \LARGE Stronger Together: \\ Air-Ground Robotic Collaboration Using Semantics}
\author{Ian D. Miller$^{1}$, Fernando Cladera$^{1}$, Trey Smith$^{2}$, Camillo Jose Taylor$^{1}$, and Vijay Kumar$^{1}$
\thanks{
This work was supported by ARL DCIST CRA W911NF-17-2-0181,
NSF grants CCR-2112665, CNS-1446592, and EEC-1941529,
ONR grants N00014-20-1-2822 and N00014-20-S-B001, and
NVIDIA.
Ian Miller acknowledges the support of a NASA Space Technology Research Fellowship.
}
\thanks{$^{1}$ Ian D. Miller, Fernando Cladera, Camillo Jose Taylor, and Vijay Kumar are with the
GRASP Lab, University of Pennsylvania, Philadelphia, PA 19104.
{\tt\footnotesize iandm@seas.upenn.edu}; {\tt\footnotesize fclad@seas.upenn.edu}; {\tt\footnotesize cjtaylor@cis.upenn.edu}; {\tt\footnotesize kumar@seas.upenn.edu}}
\thanks{$^{2}$ Trey Smith is with the NASA Ames Intelligent Robotics Group, Moffett Field, CA 94035
{\tt\footnotesize trey.smith@nasa.gov}}
}

\newcommand\copyrighttext{%
  \footnotesize \textcopyright 2022 IEEE. Personal use of this material is permitted.
  Permission from IEEE must be obtained for all other uses, in any current or future
  media, including reprinting/republishing this material for advertising or promotional
  purposes, creating new collective works, for resale or redistribution to servers or
  lists, or reuse of any copyrighted component of this work in other works.}
\newcommand\copyrightnotice{%
\begin{tikzpicture}[remember picture,overlay]
\node[anchor=south,yshift=10pt] at (current page.south) {\fbox{\parbox{\dimexpr\textwidth-\fboxsep-\fboxrule\relax}{\copyrighttext}}};
\end{tikzpicture}%
}

\begin{document}

\maketitle
\copyrightnotice
\begin{abstract}
 In this work, we present an end-to-end heterogeneous multi-robot system framework where ground robots are able to localize, plan, and navigate in a semantic map created in real time by a high-altitude quadrotor. The ground robots choose and deconflict their targets independently, without any external intervention. Moreover, they perform cross-view localization by matching their local maps with the overhead map
 using semantics. The communication backbone is opportunistic and distributed, allowing the entire system to operate with no external infrastructure aside from GPS for the quadrotor.  We extensively tested our system by performing different missions on top of our framework
 over multiple experiments in different environments.  Our ground robots travelled over 6 km autonomously
 with minimal intervention in the real world and over 96 km in simulation without interventions.
\end{abstract}

\section{Introduction}
\label{sec:introduction}

Multi-robot systems offer many advantages in terms of resilience to failure, adaptability, and the ability to complete work in parallel
 \cite{dorigo2020reflections}.
In particular, heterogeneous systems can take advantage of the specific strengths of different robots in the team to complete a task more effectively compared to homogeneous unspecialized teams~\cite{notomista2021resilient}.  
Most terrestrial mobile robots are either unmanned ground vehicles (UGVs) or aerial vehicles (UAVs), and so a number
of works have investigated synergies between these two groups~\cite{caska2014survey, alamouri2019joint, mueggler2014aerial}.

Although there has been extensive research on multi-robot systems \cite{collins2021scalable, ropero2019terra}, most of it has been in simulation. 
Simulations often offer benign conditions compared to field-robotic scenarios: many assume some level of communication between robots, robot localization in a common
frame, and the existence of low-level robot planning and navigation. 
There still exist many problems that must be solved to enable widespread real-world
operation of air-ground robot teams. 

One key application for air-ground teams is search-and-rescue.
In an urban environment, ground-level GPS may be unavailable or highly degraded.
Aerial robots can rapidly survey the scene but may not be able to fly in a highly cluttered near-ground environment, making it challenging to see details such as survivors inside of buildings.
By contrast, ground robots can easily provide these details, but are relatively slow and have a more limited sensing area.
Therefore, it is desirable to use the UAV for high-altitude scouting and as communication infrastructure.
UGVs can use UAV information to comprehend the global context and plan paths to areas that require further inspection.

\begin{figure}
    \centering
    \includegraphics[width=0.75\linewidth]{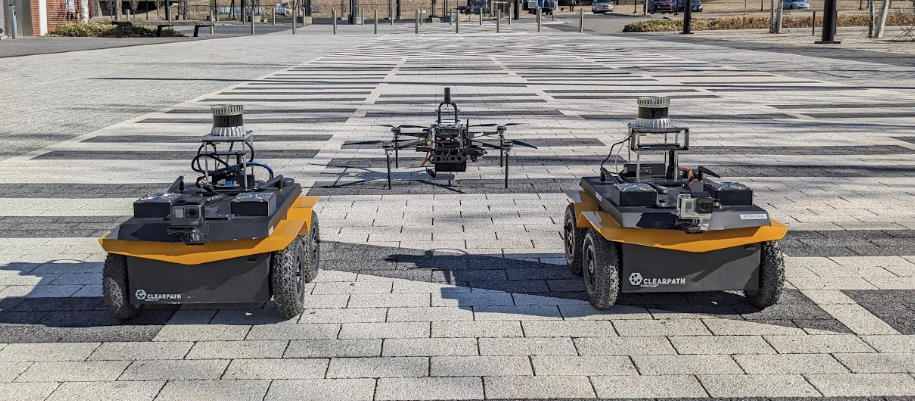}
    \caption{Robots used during our experiments, the Falcon 4 quadrotor and the Clearpath Jackals.}
    \label{fig:robots}
    \vspace{-5mm}
\end{figure}

Our work seeks to enable such applications and advance the state-of-the-art by building a system framework for multi-robot autonomy 
achieving all of the following properties:
\begin{itemize}
    \item \textbf{GPS-limited:} Only the aerial robot has GPS.  Given high-quality odometry with scale, our
        system could operate entirely without GPS.
    \item \textbf{Online:} No post-processing is required and all algorithms run in real-time onboard the robots.
    \item \textbf{Distributed:} There is no central coordinating server, and our basestation is used only for
        system monitoring.
    \item \textbf{Heterogeneous:} We integrate UAVs and UGVs, each with distinct sensors that utilize the platforms' strengths.
    \item \textbf{Intermittent Communication:} No infrastructure beyond the robots themselves is assumed.  Communication links are used opportunistically when available.
    \item \textbf{Collaborative Planning:} UGVs use UAV data to help them make intelligent decisions about obstacles and terrain traversability.
    \item \textbf{Scale:} We demonstrate the method to scale to large environments as well as large teams.
\end{itemize}

We choose cell-wise semantic object classes as a common map representation for our robot team.
Advances in deep learning have resulted in
great progress in semantic classification and detection~\cite{garciagarcia2017review}.  In addition, semantics are viewpoint-invariant, making
them an attractive basis for fusing information from different viewpoints and sensors.  In our setting, they provide contextual information about
traversability and regions of potential interest for decision-making.  Our system uses semantics for localization, 
global planning, and goal selection, weaving a class-level understanding of the world throughout the design. 
Finally, semantics are understandable by human operators and provide useful context for the operator's situational awareness.

Our contributions are as follows:
\begin{itemize}
    \item A complete integrated framework for GPS-denied ground robots and aerial robots in real-world environments 
    including localization, communication, control, and planning subsystems.
    \item Demonstration of the proposed framework on multiple missions in two real-world large-scale representative environments, and experiments with up to 10 ground robots in simulation.
    \item Public release of the aerial mapping portion of the codebase, which we call Aerial Semantic Online Ortho-Mapping (ASOOM)\footnote{\url{https://github.com/iandouglas96/asoom\_oss}}.
\end{itemize}

While all the individual components of our system may not be novel, to our knowledge, this is the first integration of a complete system that satisfies all of our desired properties while emphasizing semantic representations at each level of autonomy.

\section{Related Work}
There has been and continues to be interest in the research community around robotic collaboration, 
especially between aerial and ground systems~\cite{ding2021review}.  The roles of the air and ground
robots, as well as the expected environments, vary greatly between these works. 

\begin{table*}[t]
    \centering
    \vspace{2mm}
    \begin{tabular}{c|c|c|c|c|c|c|c|c|c}
        & Ours & \cite{qin2019autonomous} & \cite{delmerico2017active} & \cite{su2021framework} & \cite{alamouri2019joint} & \cite{yue2020collaborative} & \cite{grocholsky2006cooperative} & \cite{peterson2018online} & \cite{shah2022viking} \\
        \hline \hline                           
        GPS-limited Localization    & X & X &   &   &   & X &   &   &   \\ \hline
        Online                      & X & X & X & X &   & X & X & X & X \\ \hline
        Distributed                 & X & X &   &   &   & X & X & X &   \\ \hline
        Heterogeneous Sensing       & X &   &   &   & X &   & X & X &   \\ \hline
        Intermittent Communication  & X &   &   &   &   & X &   &   &   \\ \hline
        Exp. Scale (m)              & 200 (400 sim) & 40 & 40 & 20 & 200 & 200 & 100 & 100 & 1000 \\ \hline
        Collaborative Planning      & X & X & X &   &   &   &   & X & X \\ \hline
        Scaling (\#Air/\#Ground)    & 1/2 (10 sim) & 1/1 & 1/1 & 1/1 & 1/4 & 1/1 & 1/1 & 1/1 & 0/1 \\
    \end{tabular}
    \caption{Comparisons of related works}
    \label{tab:related_work}
    \vspace{-7mm}
\end{table*}

\subsection{Collaboration}
We compare some of the most similar works to ours in Table~\ref{tab:related_work} using the metrics identified in Sec.~\ref{sec:introduction}.
The scales of the experiments are measured as the approximate square root of the total area of the experiment region.
Works which did not explicitly state the size of their experiments are given sizes approximated from the provided maps.
Many authors do not address the problems of localization or planning, or only address the single-robot case.
However, we argue that while this more specific work is valuable, it is nevertheless important to demonstrate the integration of these modules in a unified system.

Several authors have also sought to develop such a system.
Qin \etal~\cite{qin2019autonomous} present a UAV/UGV team with the objective of exploring indoor environments.
The UGV first builds a coarse map, and the UAV follows with a refining pass, both performing active exploration.
However, this is a sequential process, not simultaneous, so communication is not addressed.
In addition, both robots are equipped with LiDAR sensors and have relatively similar viewpoints, and experiments are small-scale at the size of a room or small office space.
The authors of \cite{yue2020collaborative} implement a similar LiDAR-based system, but incorporate semantics and consider simultaneous operation with imperfect communication.
They demonstrate their system with a UAV/UGV team and a UGV/UGV team, but the viewpoints of the agents are similar, and the authors do not discuss planning at all, focusing instead on mapping.
Due to the robots' relatively similar viewpoints, the benefits of the heterogeneity of the team are not fully realized.

Other works address this problem, but assume GPS availability for relative localization.
The ANKommEn project~\cite{alamouri2019joint, alamouri2021development} uses UAVs and UGVs to build multi-modal maps.
GPS is used as the primary means of localization, though the authors perform map alignment for further refinement as a post-processing step.
Their aerial mapping system is similar to ours, but does not generate semantic maps.
The authors of \cite{grocholsky2006cooperative} exploit the heterogeneity of the air/ground team to localize targets.  However, localization and navigation
are performed independently using GPS, and obstacle avoidance is not considered.
In a similar work \cite{su2021framework}, the authors use a UAV to 
detect targets from the air and dispatch the ground robot to their locations.
Here obstacle avoidance is considered for ground robots, but only independently without using aerial data.

Many authors have recognized the advantages
of exploiting aerial data for ground robot planning, since many obstacles and/or challenging terrain can be seen from the air.
Peterson \etal~\cite{peterson2018online} build an orthomap using a UAV and classify the terrain in the
map using classical computer vision methods, which they then use to plan safe paths for a UGV.
They perform outdoor experiments, and emphasize that their system
is designed to run in real-time, similar to ours.  However, relative localization is performed using GPS.
Fedorenko \etal \cite{fedorenko2018global} take a similar approach, but instead build a high-resolution point cloud
of the region and use the topography data to establish traversability. 
Recently, the power of neural networks has enabled methods that perform semantic segmentation to classify terrain and plan safe paths \cite{delmerico2017active}.
Other authors take a different approach, training
a neural network to reason about local traversability while being informed by satellite data \cite{shah2022viking}, but do not extend the method to real-time aerial maps or multiple UGVs.

Although GPS is readily available for aerial robots flying over buildings, it is often noisy and difficult to use for ground robots in cluttered urban environments.
Some authors, recognizing this, have worked on GPS-constrained relative localization schemes.
Vandapel \etal~\cite{vandapel2006unmanned} address this problem in the outdoor setting by creating a high-fidelity LiDAR map
from the air.  They then match ground scans against this map, as well as computing traversability metrics.
However, this approach relies on a very high-resolution aerial map built in advance of ground robot operation.
Kaslin \etal~\cite{kaslin2016collaborative} similarly rely on geometric features for cross-registration, but
build an aerial map using photogrammetry from a quadrotor.  They then match the elevation maps between the
aerial and ground robots using a template matching approach with a particle filter.

We use the localization algorithm presented in our prior work \cite{miller2021any} with some modifications, but in this work
demonstrate it operating with dynamically updated maps, running on constrained robot hardware in-the-loop with planning and control,
and in the context of larger communication, planning, and coordination systems.

As shown in Table \ref{tab:related_work}, while our work builds on a number of existing methods, many of these methods have not previously been shown to work effectively in-the-loop, onboard robots, in real-world communication environments, at scale, with multiple robots, with minimal GPS.

\subsection{Communication}
Communication is itself a large research area that our work builds on.
Some authors have investigated the use of aerial robots to provide network infrastructure to other
ground or aerial robots that work to perform a task \cite{mox2020mobile, nouri20213d}, with particular
emphasis on determining the optimal placement of these aerial relays.  Others have designed networks that incorporate
data caching in aerial robots~\cite{cheng2018air}. 
Recent works have focused on multi-agent communication in harsh real-world scenarios, particularly without a preexisting communication infrastructure. Ginting \etal~\cite{ginting2021chord} propose using the ROS2 Data Distribution System to transmit messages between heterogeneous agents in subterranean environments using commercial mesh networks. Other approaches use low-bandwidth communication channels such as LoRa~\cite{huang2019duckiefloat}.

Our system is built from the ground up with the perspective that communication links will only be available sporadically and that robots should take advantage of communication links opportunistically.
\section{Method}
\begin{figure}
    \centering
    \includegraphics[width=\linewidth]{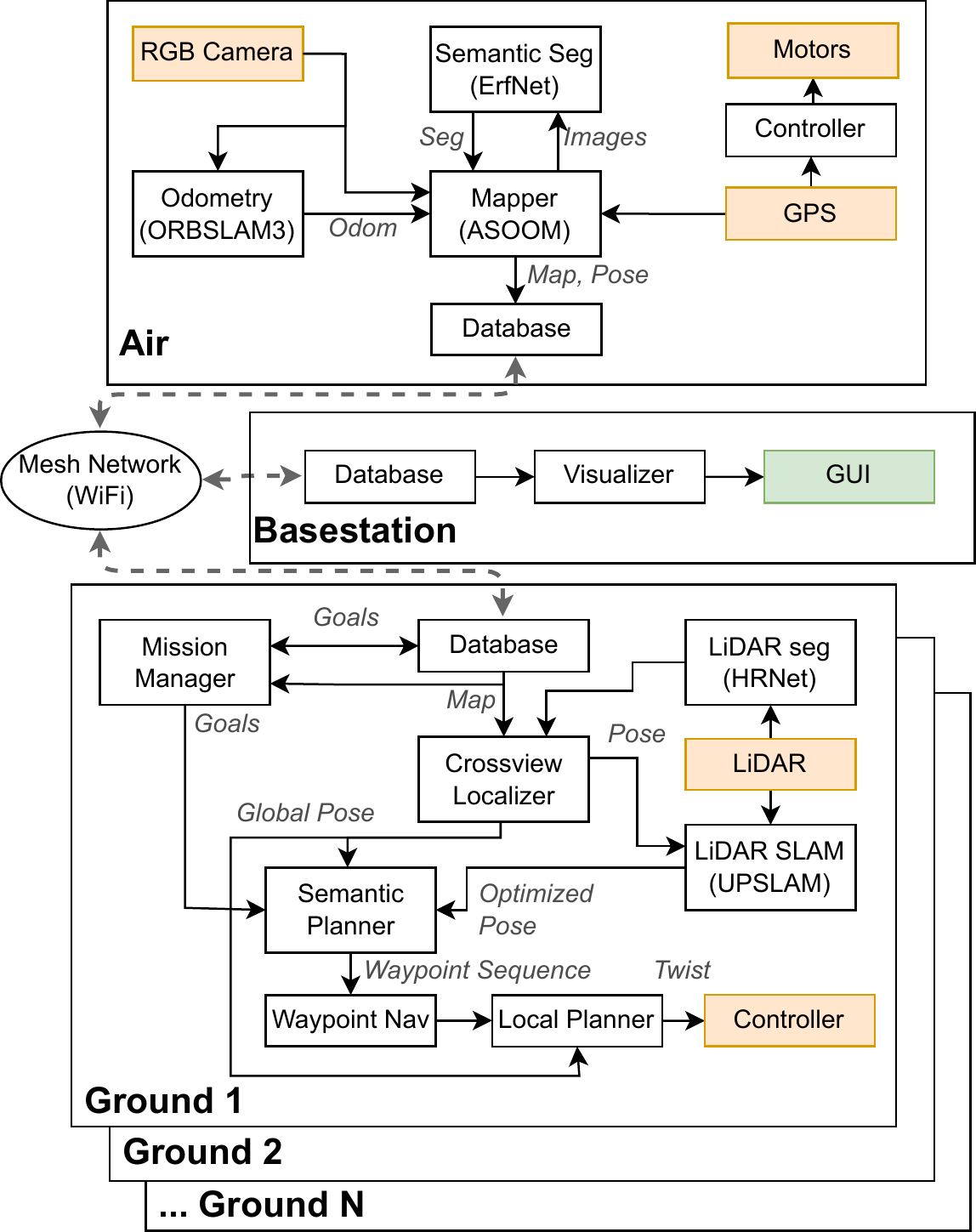}
    \caption{The full system architecture.  Orange indicates sensor and actuation interfaces, and green indicates output visualization tools.}
    \label{fig:block_diag}
    \vspace{-5mm}
\end{figure}

Our experimental system consists of two Clearpath Jackals and a single custom-built quadrotor (the Falcon 4).
We additionally have a single computer acting as a base station for visualization.  A high-level overview
of the system architecture is shown in Fig.~\ref{fig:block_diag}, and pictures of our robots
in Fig.~\ref{fig:robots}.

\subsection{Hardware}

\subsubsection{Quadrotor}
We use the Falcon 4~\cite{liu2022large} as our aerial platform for the experiment.  The primary
compute on the robot is an Intel NUC 10 with an Intel i7-10710U processor. 
We employ an Open Vision Computer 3 (OVC)~\cite{quigley2019ovc} global shutter RGB camera, a
PX4-based flight controller, and a ZED-F9P GPS as the sensor suite.

\subsubsection{Ground Robots}
Our ground robots are based on the Clearpath Jackal platform. We
modified the Jackals' compute platform to feature an AMD Ryzen 3600 CPU
and an NVIDIA GTX 1650 GPU. We use an Ouster OS1-64 LiDAR as our
perception sensor, and for some experiments an OVC as well for object detection.
The ground robots are also equipped with GPS, which is used to record ground truth location for comparison purposes only.

\subsection{Aerial Mapping}
Our Aerial Semantic Online Ortho-Mapper (ASOOM) is inspired by the work of Hinzmann \etal~\cite{hinzmann2017mapping} and similar to OpenREALM~\cite{kern2020openrealm}. Similar to these methods,
ASOOM takes as input odometry from an external source, in our case ORBSLAM3~\cite{campos2021orb} in
pure monocular mode without global loop closure search. 
Once the robot has moved a threshold distance we create a new keyframe.  Sequential
keyframes are rectified and dense stereo is computed using block matching.  The densified stereo
is then projected into a point cloud, and each point is assigned to its containing cell in the $xy$ plane.  The elevation
is tracked for each cell as a cumulative average, and, following Hinzmann, we assign each cell the color
and class from the image with the corresponding pixel closest to the image center.

\begin{figure}
    \centering
    \vspace{2mm}
    \includegraphics[width=0.8\linewidth]{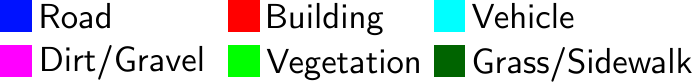}
    \caption{Semantic classes for our experiments, as well as colors used for the rest of this work.}
    \label{fig:classes}
    \vspace{-6mm}
\end{figure}

Unlike other methods, however, we also employ a pose graph back-end built on GTSAM~\cite{dellaert2012factor}.
The pose graph consists of transform edges between keyframes, determined by ORBSLAM3, and GPS measurements.
The optimizer is also allowed to determine the overall scale of the map, thereby resolving the scale
ambiguity of monocular odometry using GPS.  Additionally, each keyframe is segmented using ErfNet~\cite{romera2018erfnet}
on the CPU with classes shown in Fig.~\ref{fig:classes}.

ASOOM runs easily in real-time on our aerial robot using the onboard Intel NUC. We have also run the entire ORBSLAM3, ASOOM, and segmentation pipeline on an NVIDIA Xavier NX; however, our implementation on this platform can only support a frame rate of 10Hz since ORBSLAM3 takes at least 100ms to process a frame.

\subsection{Localization}
Our ground localization system is built on top of our prior work~\cite{miller2021any}.  In short, the LiDAR scan
is segmented using HRNets~\cite{wang2020deep} and projected onto a top-down polar semantic map.  This map is 
then compared with the aerial semantic ortho-map inside of a particle filter.  We perform LiDAR odometry using 
UPSLAM~\cite{cowley2021upslam}. The output from the semantic localizer is fed back into UPSLAM in the form 
of pose graph constraints, grounding the final estimation to the global frame.

We make several modifications to the localizer for this work.  First, the loop closure search with UPSLAM is
disabled for the region investigation experiments, both to save computational load and to avoid large jumps in
odometry.  Second, since the aerial map is incomplete, we must handle comparisons with unknown 
regions.  When computing the matching cost, if there is no known class for a given cell in the aerial
map, we assign a fixed cost.

In the experiments in this work, we assume that the ground robots begin at a fixed location and orientation
in the map.  
This is not a hard requirement of our system: our robots could localize with respect to the aerial map after the particle filter converges to the initial position. 
However, a limitation of our system is that the UAV must have mapped the region where a UGV is when its localizer initializes, even if its exact location is unknown.
In practice, we believe this to be a reasonable assumption, as all robots typically start from some known staging location.

We emphasize that a key novelty over prior work is that both air and ground maps are built in real-time, as opposed to assuming their sequential construction.

\subsection{Communication}
\subsubsection{Physical and Network Layers}
For our communication hardware, we employ commercial off-the-shelf IEEE 802.11g wireless adapters operating in ad-hoc mode. 
On top of our ad-hoc wireless network we run Babel~\cite{RFC6126}, which handles routing in the mesh network.

\subsubsection{Application Layer}
The main goal of our system is to reliably transmit information between robots when they are in communication range. Additionally, we want our robots to act as \emph{data mules}: for example, if the aerial robot receives information from one of the ground robots regarding its goals, it is useful to relay this information to the other ground robot so that it can adjust its own goals accordingly. 

We use the distributed database developed in our earlier work~\cite{miller2020mine}.
In short, each robot maintains a database of its most recent understanding of the data from each robot.
When two robots come into communication range, they update each other with the most recent information available to each. This distributed, gossip-based scheme allows us to distribute relevant information to all agents in the team.

\subsection{Ground Navigation}
Our ground navigation system is composed of 3 primary components: a mission manager which defines the mission specification, a global semantic planner,
and a local waypoint tracker.  For local navigation and obstacle avoidance, we use the standard ROS
\texttt{move\_base} navigation stack.

\subsubsection{Global Semantic Planner}
By planning in the aerial map, we enable the ground robots to plan long-range paths through regions they themselves have
never visited.  We assume that all regions of class \texttt{road} and \texttt{dirt/gravel} are
traversable, and maintain a roadmap consisting of a graph of locations and traversable edges.  When a new map is received,
holes are filled with a morphological close operation.  A graph is then
built in a similar fashion to a probabilistic roadmap (PRM), but instead of randomly sampling we compute
for each cell in the 2D map the set of cells within a radius $R$ which have an unobstructed straight edge to the node.
We call this the visibility map.  In addition, we compute the distance transform, which encodes the distance
from each cell to the nearest obstacle. 
The graph is constructed by iteratively adding nodes corresponding to cells with the largest distance value that are not visible to any node in the current graph.
Edges are added to neighboring nodes, and if there
is another node with overlapping visibility and no path between them, a new node is added in the overlapping
region.  The weight of the edge $(u, v)$ is given by the heuristic
\begin{align}
    W_{u,v} = \lambda\|u - v\|_2 + \left[\min{d(u, v)}^2 + \sqrt{\min{d(u, v)}}\right]
\end{align}
where $d(u, v)$ is the set of distances to the nearest obstacle for all points between $u$ and $v$ and
$\lambda$ controls the distance/safety tradeoff.

\begin{figure}
    \vspace{2mm}
    \centering
    \includegraphics[width=0.85\linewidth]{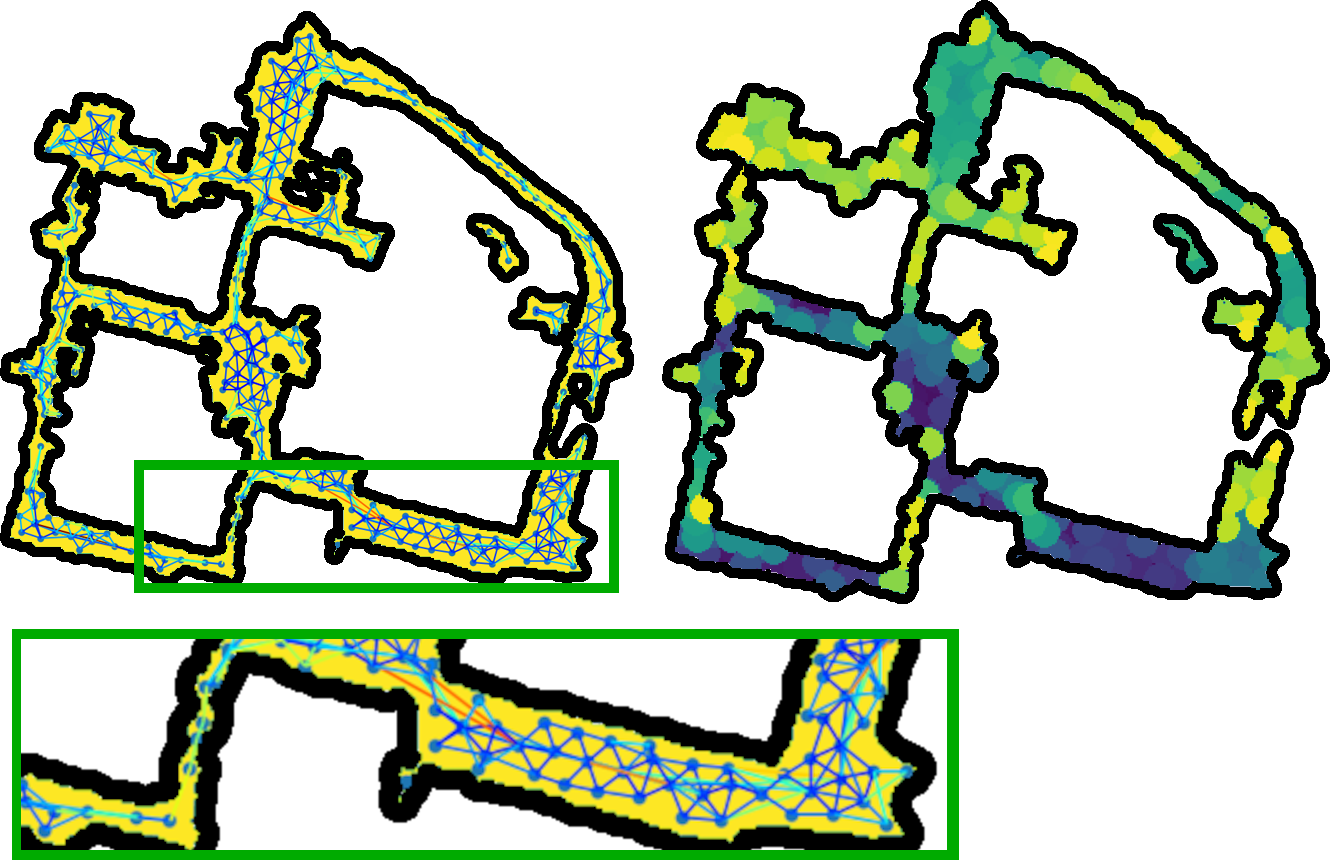}
    \caption{\underline{Left}: Roadmap overlaid onto the map.  The edge colors correspond to weights, with lighter colors indicating greater weight. \underline{Right}: Visibility map, where different colors indicate visibility to different nodes.}
    \label{fig:roadmap}
    \vspace{-5mm}
\end{figure}

This method allows us to incrementally build roadmaps on each ground robot, updating them as map updates
arrive from the aerial robot.  Examples of the roadmap and visibility map are shown in Fig.~\ref{fig:roadmap}.
The node density is controlled by selecting a value for the parameter $R$.  To plan between two points, we use the visibility map
to first look up the nodes visible to the start and end, perform a graph search, and finally prune the resulting path.

\subsubsection{Local Waypoint Tracker}
Given a set of waypoints, either from the global semantic planner or a manual specification, we generate goals 
for the local planner.  For each keyframe depth panorama from UPSLAM, we fit a ground plane using RANSAC and
select all points within a threshold of this plane.  We project a ray to each of these points which clear
space in the obstacle map.  All other points are added as obstacles.  We then select a goal point in the
direction of the next waypoint in known space, and for each cell in a region around that point compute the 
distance to the nearest obstacle.  The point with the greatest distance becomes the local goal.  If the 
region is entirely occupied, the planner first tries to backtrack to the last waypoint.  If that fails, 
the path is cancelled. 

A simple local obstacle avoidance system also analyzes the instantaneous LiDAR scan and treats all points above a specified height as obstacles.

\section{Results}

We performed two sets of real-world experiments in different locations with different objectives to emphasize the versatility
and capability of our proposed system.
We additionally ran experiments in simulation with larger robot teams to investigate effects of scale and compare to ground truth localization.

\subsection{Target Mapping}
\begin{figure}
    \vspace{2mm}
    \centering
    \includegraphics[width=0.6\linewidth]{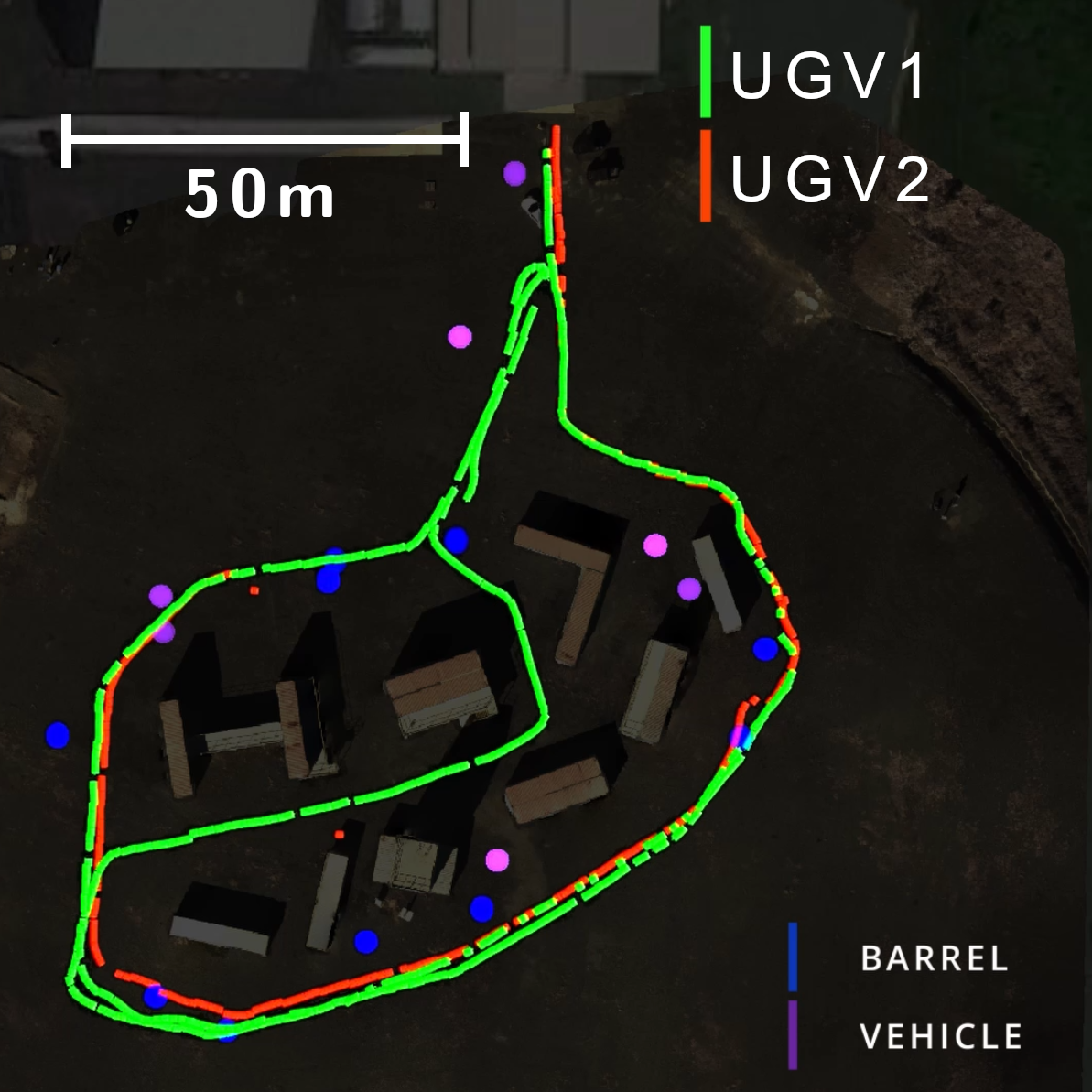}
    \caption{Trajectories and object detections as seen by the base station after one target mapping experiment.}
    \label{fig:gqtraj}
    \vspace{-5mm}
\end{figure}

For our first mission, the robots had the objective of locating a set of target objects placed at unknown locations in the scene.
This task could be thought of as localizing people in a location where the operator has access to satellite imagery and knows the roads to search.
In advance, we flew our quadrotor over the area at an altitude of 35~m
and built an offline georeferenced photogrammetric map using OpenDroneMap which can be seen in Fig.~\ref{fig:gqtraj}.
The map was then manually semantically labelled. The ground robots segmented their
LiDAR scans for localization by detecting points of high gradient and flat regions instead of using HRNet, effectively creating two classes: \texttt{obstacle} and \texttt{not obstacle}.

The Jackals were also equipped with an RGB camera to detect barrels and cars set up throughout the environment
using YOLOv3~\cite{redmon2018yolov3}.  We used an Extended Kalman Filter to filter the object locations over time.  The robots followed
a predetermined set of waypoints specified manually in the semantic map.

We completed 4 runs of this experiment, and the robot trajectories
and object detections for one experiment as received by the base station are shown in Fig.~\ref{fig:gqtraj}.
Between the 2 ground robots and all 4 runs, the Jackals autonomously travelled over 4 km. This distance included
instances of obstacle avoidance of the barrels and vehicles with only several manual interventions in total.

The distributed database was able to successfully relay data back to the base station and
the quadrotor.
This experiment demonstrated tasking of our robots to perform missions based on the overhead map acquired by our aerial robot. 

\subsection{Region Investigation}
For our second mission, the objective of the ground robots was to inspect targets of interest that were initially detected in the aerial imagery.
This task is similar to the problem of investigating cars detected from the air to search for associated survivors.
These regions of interest (ROIs) were selected by clustering
vehicle detections in the aerial map.  Robots put their list of current and prior targets in the database
and, when selecting a new goal, chose an ROI that had not yet been claimed in order to avoid redundancy.
The goal point was selected to be the point in the ROI furthest from an obstacle.

For this mission, the aerial map was built in real-time using ASOOM and was continuously updated as the ground robots performed their tasks.
The quadrotor flew missions specified by GPS waypoints 
at an altitude of 40~m and acted as both a provider of semantic maps to the ground robots and a communication relay for data.
The ground robots independently selected goal regions and planned paths to those regions using the aerial map, localizing themselves in the map in real-time.  

\begin{figure}
    \vspace{2mm}
    \centering
    \includegraphics[width=\linewidth]{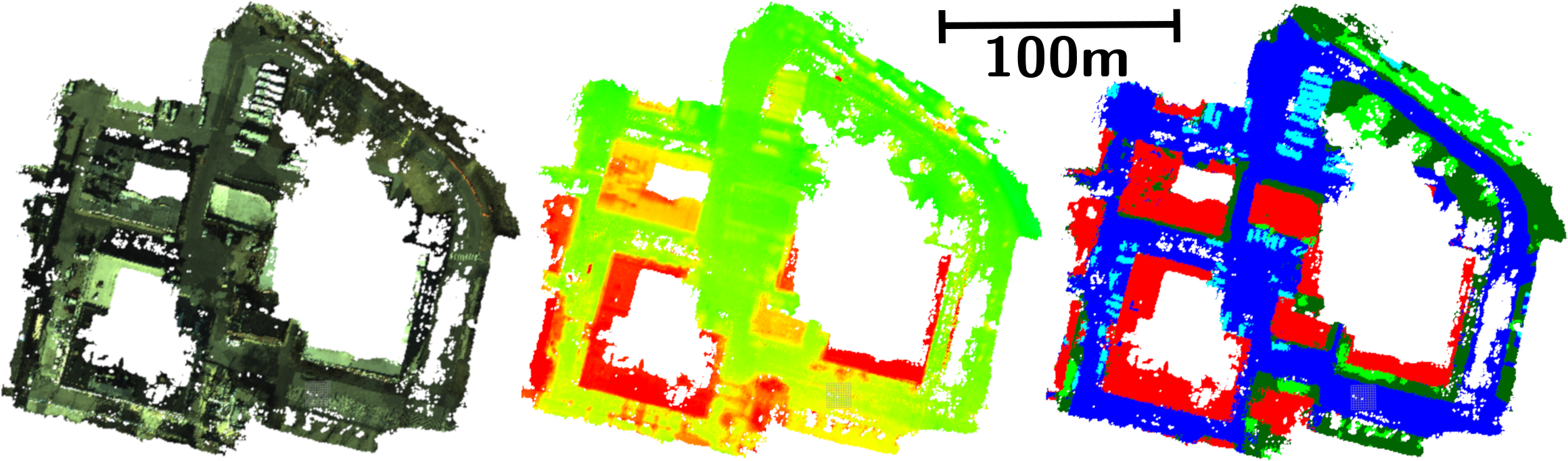}
    \caption{\underline{Left to right}: RGB, elevation, and semantic maps built in real-time by ASOOM on-board our UAV.}
    \label{fig:asoom_maps}
    \vspace{-5mm}
\end{figure}

We conducted 4 experiments around the Pennovation campus.  Throughout the experiments, a total of 26 target regions
were automatically identified, and 22 of these were visited by at least one robot.
Together, the ground robots autonomously
traversed over 2.5 km.
Fig.~\ref{fig:asoom_maps} shows a map built in real-time by ASOOM on the quadrotor in one experiment,
incorporating RGB color, elevation, and semantic layers.

\begin{figure}[b]
    \vspace{-5mm}
    \centering
    \includegraphics[width=0.9\linewidth]{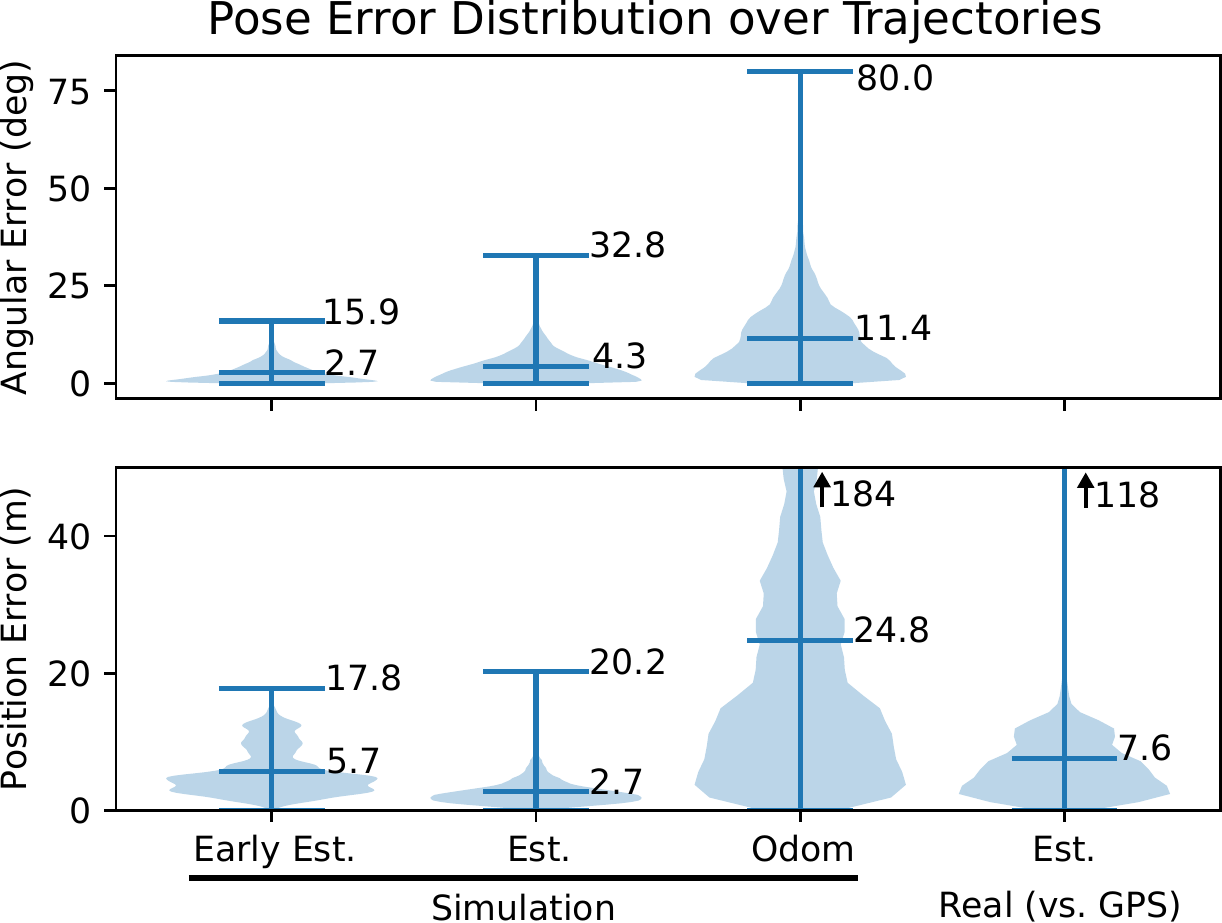}
    \caption{Pose errors over all robots over all trajectories.  Position error for real experiments is with respect to GPS. Means and maxima are labelled.}
    \label{fig:pose_errors}
\end{figure}

\begin{figure}
    \vspace{2mm}
    \centering
    \includegraphics[width=\linewidth]{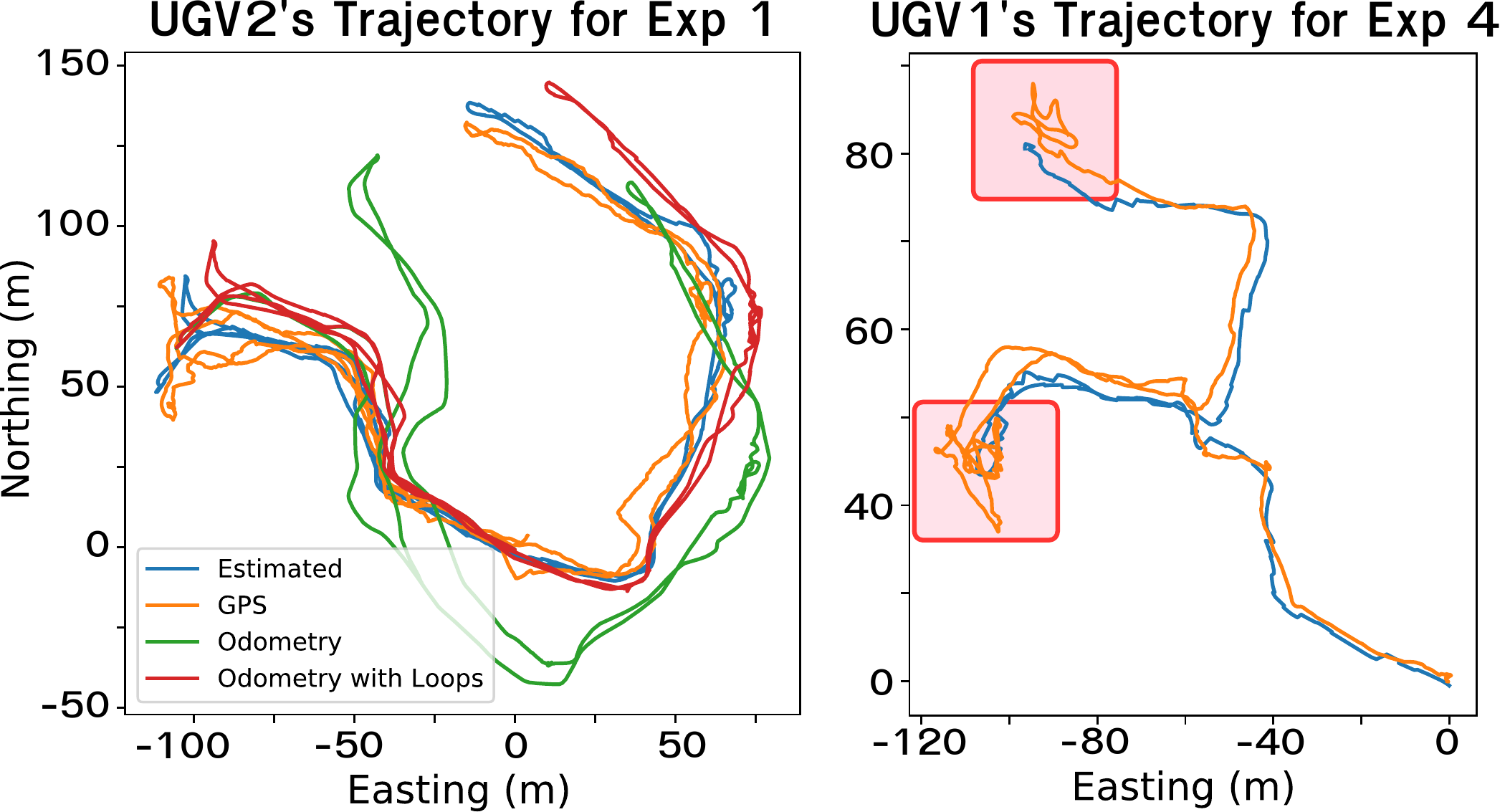}
    \caption{GPS and estimated trajectories for two robots on different experiments.  The highlighted red regions in the right plot indicate areas where the robot was stationary, but there was significant GPS noise.}
    \label{fig:example_trajectories}
    \vspace{-4mm}
\end{figure}

Fig.~\ref{fig:example_trajectories} shows representative trajectories. 
We show GPS here only for comparison purposes; it was not used in the experiment itself.
Note, in particular, the motion of the GPS estimate highlighted in the red boxes.
During these times, the robot was completely stationary, but the GPS estimate
migrated 5-10 meters.
We conclude that, except for the two outlier cases, our localizer performance was strong enough that its error was below the detection limit for this GPS.  In addition, we see in Fig.~\ref{fig:example_trajectories} that when running UPSLAM without loop closures,
the accuracy is insufficient for planning and drifts significantly.  Even loop closures, while they
help, cannot sufficiently globally localize the robot within the road width, which we need to effectively navigate in the global map.
Larger errors become increasingly difficult for the local planner to recover from, especially in denser urban environments.
We conclude that our semantic
localizer is an essential component of the global navigation system.

Fig.~\ref{fig:pose_errors} shows the distribution of position error over all trajectories and robots compared to GPS in the rightmost violin plot. 
Due to the aforementioned GPS drift, it is difficult to tell how much error is due to the GPS, and how much due to our localizer.
In our experience, most large localizer errors occurred when the ground robots sat still
for a long time and UPSLAM drifted in its orientation estimation.  The orientation confidence was set very high,
so the semantic localizer drifted as well, and when the robot began moving again the estimate rapidly diverged.
For later experiments the confidence was decreased, mitigating this problem.

In these experiments the planner performed well, avoiding obstacles such as poles in its path.
Manual intervention was typically required a few times for each robot over
the course of an experiment.  These interventions were usually due to curbs or benches, which our simple obstacle
detector did not see.
Additionally, the aforementioned yaw drift sometimes resulted in the waypoints being poorly aligned with the map, causing poor planner performance.

\begin{figure}
    \centering
    \includegraphics[width=0.9\linewidth]{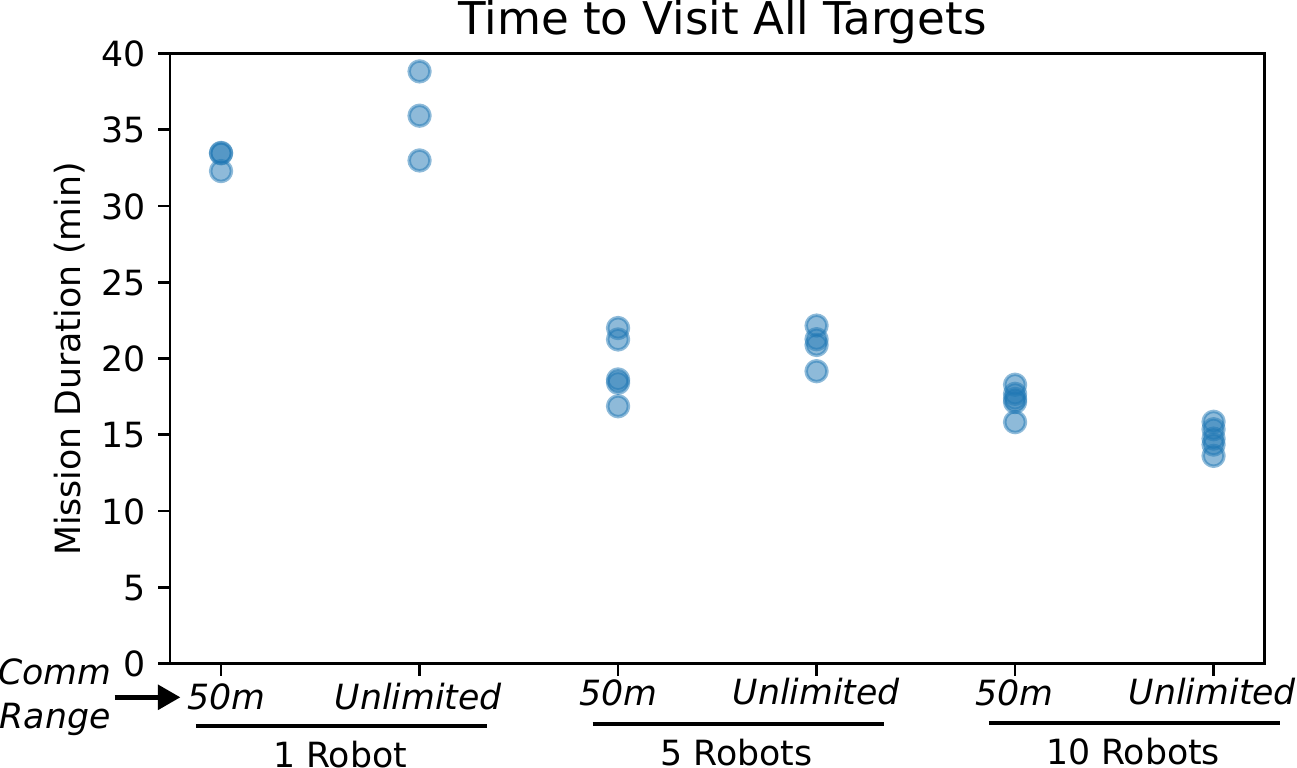}
    \caption{The times required to visit all targets with varying team sizes and communication ranges.}
    \label{fig:mission_durations}
    \vspace{-8mm}
\end{figure}

\begin{figure}
    \vspace{2mm}
    \centering
    \includegraphics[width=0.75\linewidth]{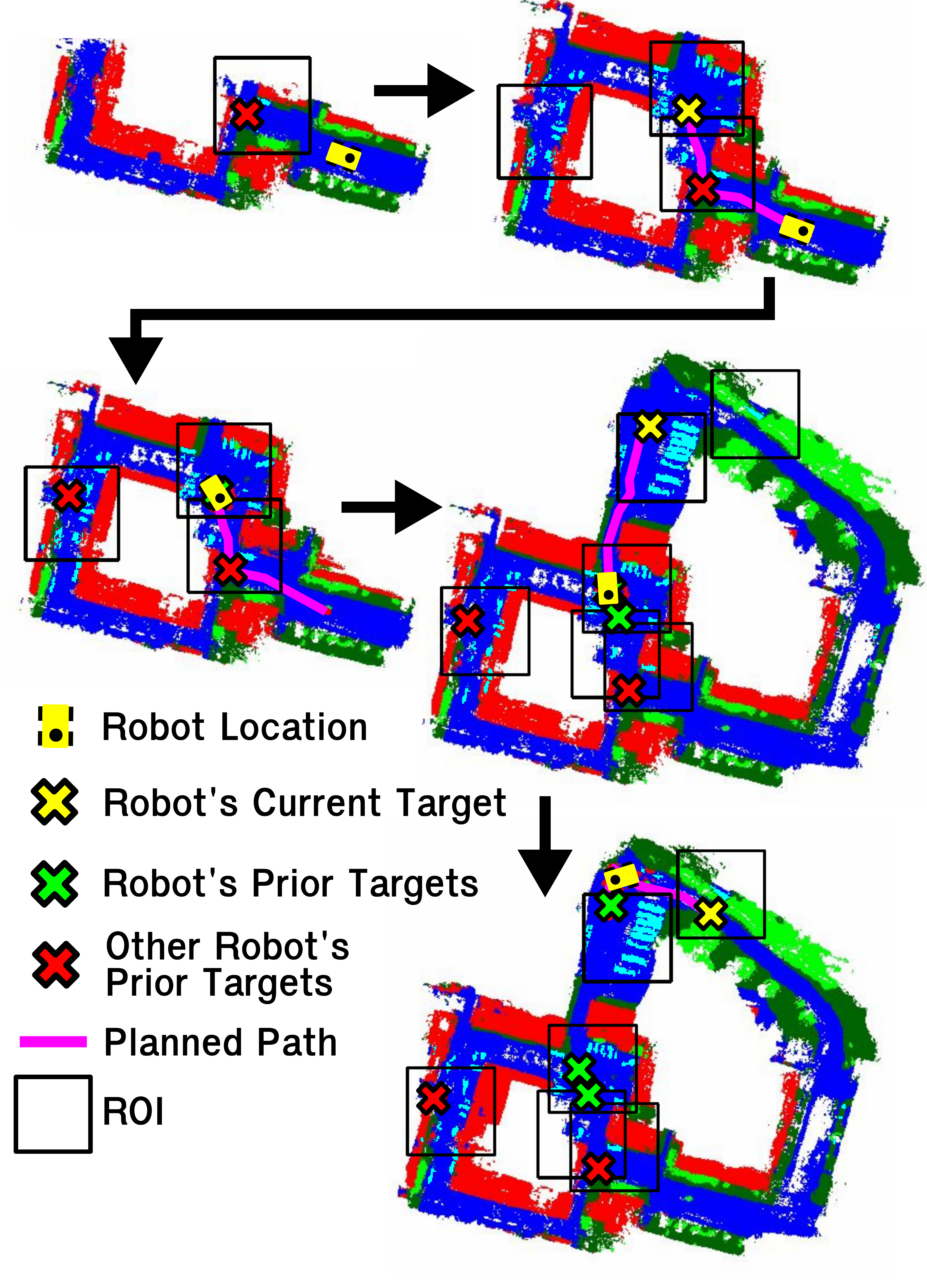}
    \caption{A portion of a single region investigation experiment from the perspective from one of the UGVs.}
    \label{fig:scenario1}
    \vspace{-5mm}
\end{figure}

The distributed database, networking, aerial mapping, and goal generation subsystems demonstrated their utility
in enabling complex high-level behavior.
Fig.~\ref{fig:scenario1} shows one scenario from an experiment that demonstrates an interaction between all robots.
Initially the UGV only saw a single ROI, but it was already claimed by the other robot and no action was taken.  Next, a map update from the UAV revealed a new ROI, so a goal was selected and planned to
using the aerial map.  At this point the robot again waited for an un-claimed ROI, which was revealed by 
new exploration by the quadrotor.  Meanwhile, the UAV relayed this information back to the
base station where it was visualized.  These sorts of interactions happened many times in our experiments
and show the behavioral complexity achieved by our system.
We observed that the performance of our system was dependent on the choice of waypoints for the quadrotor, as poor waypoints led to UGVs doing nothing for long periods without new data to act on.
This is a limitation of our system, and active navigation for the UAV is an important area for further work.

\begin{figure}
    \centering
    \includegraphics[width=0.9\linewidth]{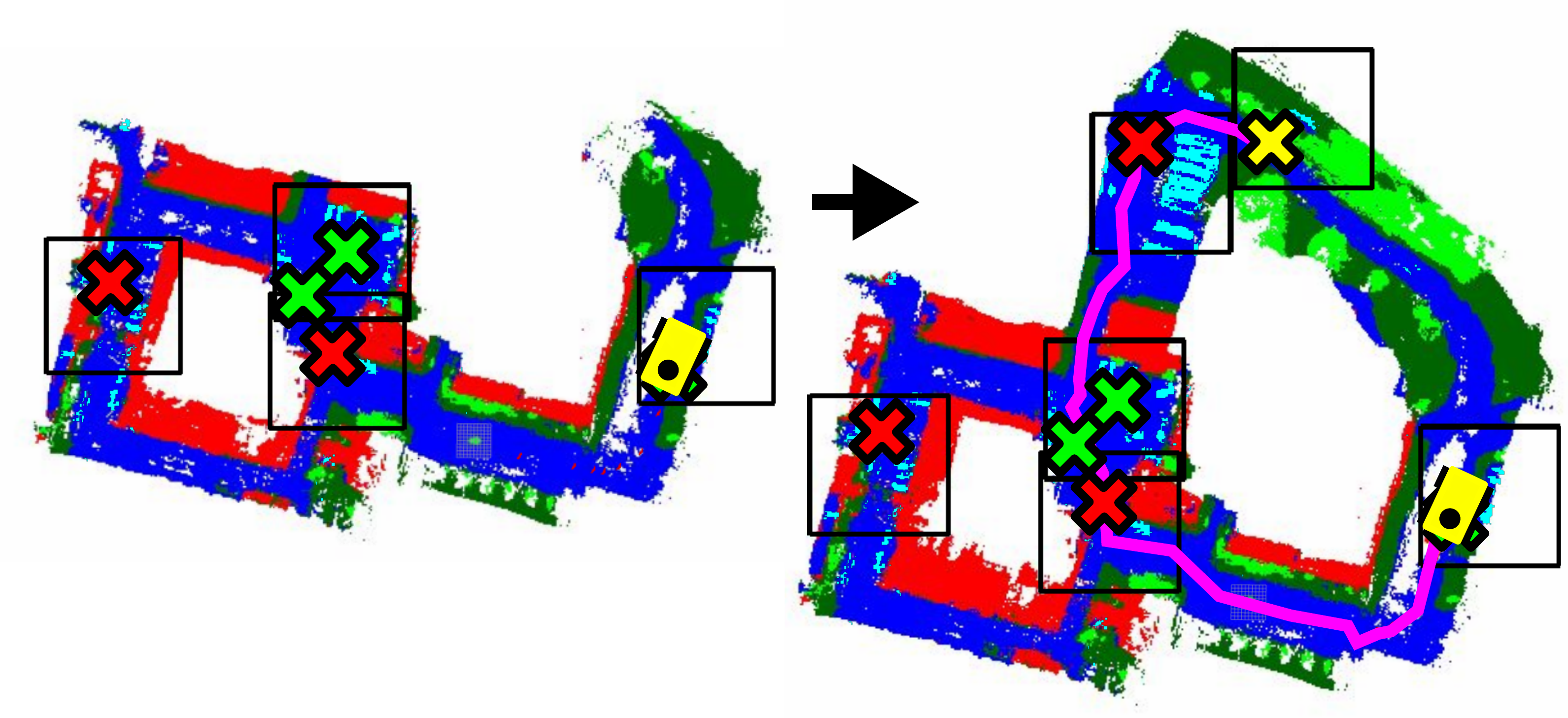}
    \caption{Another portion of a region investigation experiment from one UGV's perspective.}
    \label{fig:scenario2}
    \vspace{-7mm}
\end{figure}

In Fig.~\ref{fig:scenario2} we see the UGV receiving both a map update as well as information 
about the other robot's goals from the quadrotor, enabling the UGV to make an informed decision
about which ROI to travel to next.
This experiment demonstrated that a robot team can be tasked with a high-level objective such as inspecting all cars in an area and can then carry out all of the detection, localization, and planning processes automatically without further human intervention.

\subsection{Simulation Experiments}

\begin{figure}
    \centering
    \vspace{2mm}
    \includegraphics[width=0.85\linewidth]{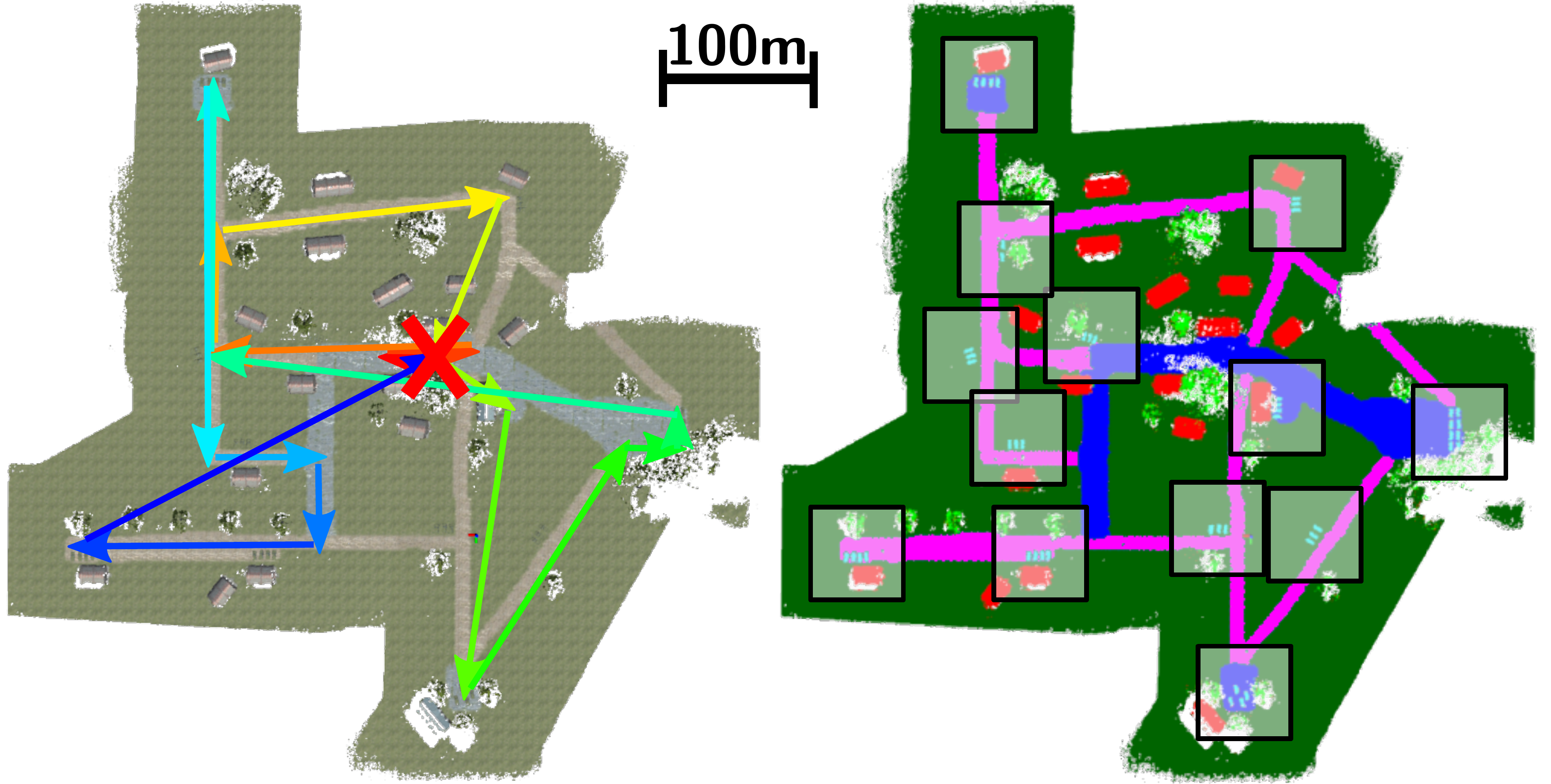}
    \caption{Color and semantic maps built by ASOOM of the simulation environment.  Targets (cars) are highlighted in the semantic map.  The red X shows the starting position of the robots and the arrows the UAV waypoint path.}
    \label{fig:sim_env}
    \vspace{-7mm}
\end{figure}

We additionally performed 26 experiments in simulation, with team sizes ranging from 1 to 10 UGVs, consisting of over 96~km travelled between all the UGVs.
The system was completely autonomous, with no manual interventions.
The UAV performed a fixed waypoint trajectory at 60~m altitude.
The full region investigation stack was run, with the exception of odometry, and with the additional ability for UGVs to inform other robots that they failed to reach their target goal.
In place of UPSLAM on the UGVs, to limit computational load, we simulated odometry by integrating the ground truth motion and adding Gaussian noise at each timestep. 
This was then fed into the semantic localizer.
We used ground truth semantic segmentation.
ASOOM was given perfect odometry from the simulated UAV in lieu of GPS, and communication was simulated by allowing all robots to communicate with other robots within a fixed distance on the $xy$ plane.
The initial UGV position estimates were up to 16~m away from ground truth.
The environment is shown in Fig.~\ref{fig:sim_env}.

The pose error distribution is shown in Fig.~\ref{fig:pose_errors}.
The early estimates are from the first minute after ground robot localization, and therefore have greater error than the rest of the run due to the localizer still converging from imperfect initialization.
On average, our localizer keeps the error below 3~m after convergence, as opposed to over 24~m when simply integrating the noisy odometry.
Across the 26 experiments, all 13 targets were reached each time with one exception.
The one exception was due to a relatively large localization error, leading the UGV to believe that it had visited the region when it had in fact not.
For the 25 completely successful experiments, we measured the time to visit all targets, beginning with the start of the UAV trajectory.
These results are shown in Fig.~\ref{fig:mission_durations}.
Note that increasing the number of robots does decrease the mission duration, although there is a saturation point.
In addition, the communication range has a minimal effect on performance, and in some cases the shorter range leads to robots closer to a goal visiting it, thereby being more efficient.

We conclude that our method can effectively scale to larger robot teams, though increasing the number of ground robots yields diminishing returns. 
To efficiently use more UGVs, it is necessary to add more UAVs to locate enough regions of interest to provide all UGVs with useful targets.
\section{Conclusion}
We have presented an integrated system for air-ground collaboration using semantic maps as a common representation
between robots, and demonstrated its efficacy with multiple field environments and mission specifications simulating problems encountered in disaster scenarios.
These experiments comprise
over 6~km of distance travelled in the real-world, and over 96~km in simulation.
UGVs operate in GPS-denied conditions, localize using the UAV's map, and globally plan using aerial
information while communicating opportunistically.
Our distributed database and semantic localizer performed very well, and we believe that this distributed architecture for UAVs and UGVs effectively exploits the strengths of both systems.
Most of our problems stemmed from the local planner, and so we conclude that highly robust local obstacle mapping and avoidance systems are key, especially in real-world environments.
In addition, we see a need for effective, scalable systems for the deployment and management of large robot teams in the field, as this was a challenge even with relatively few robots.
For our future work, we hope to improve the local planner and investigate ways that ground
information can inform other ground robots, especially with regard to planning and terrain analysis.

\section{Acknowledgements}
The authors would like to thank Jake Welde, Sahachar Reddy Tippana, Lishuo Pan, Alex Zhou, Xu Liu, and Chao Qu for
their help with field experiments, as well as the rest of Kumar Robotics.  We would also like to thank 
Drs.~Ethan Stump and Jonathan Fink and the ARL team for
their support, especially with field experiments.

\bibliographystyle{IEEEtran}
\bibliography{IEEEabrv,bib.bib}

\begin{thebibliography}{10}
\providecommand{\url}[1]{#1}
\csname url@rmstyle\endcsname
\providecommand{\newblock}{\relax}
\providecommand{\bibinfo}[2]{#2}
\providecommand\BIBentrySTDinterwordspacing{\spaceskip=0pt\relax}
\providecommand\BIBentryALTinterwordstretchfactor{4}
\providecommand\BIBentryALTinterwordspacing{\spaceskip=\fontdimen2\font plus
\BIBentryALTinterwordstretchfactor\fontdimen3\font minus
  \fontdimen4\font\relax}
\providecommand\BIBforeignlanguage[2]{{%
\expandafter\ifx\csname l@#1\endcsname\relax
\typeout{** WARNING: IEEEtran.bst: No hyphenation pattern has been}%
\typeout{** loaded for the language `#1'. Using the pattern for}%
\typeout{** the default language instead.}%
\else
\language=\csname l@#1\endcsname
\fi
#2}}

\bibitem{dorigo2020reflections}
M.~Dorigo, G.~Theraulaz, and V.~Trianni, ``Reflections on the future of swarm
  robotics,'' \emph{Science Robotics}, vol.~5, no.~49, p. eabe4385, 2020.

\bibitem{notomista2021resilient}
G.~Notomista \emph{et~al.}, ``A resilient and energy-aware task allocation
  framework for heterogeneous multirobot systems,'' \emph{IEEE Transactions on
  Robotics}, 2021.

\bibitem{caska2014survey}
S.~{\c{C}}a{\c{s}}ka and A.~Gayretli, ``A survey of uav/ugv collaborative
  systems,'' \emph{CIE44\&IMSS}, vol.~14, pp. 453--463, 2014.

\bibitem{alamouri2019joint}
A.~Alamouri \emph{et~al.}, ``The joint research project ankommen--exploration
  using automated uav and ugv,'' \emph{The International Archives of
  Photogrammetry, Remote Sensing and Spatial Information Sciences}, vol.~42,
  pp. 165--172, 2019.

\bibitem{mueggler2014aerial}
E.~Mueggler, M.~Faessler, F.~Fontana, and D.~Scaramuzza, ``Aerial-guided
  navigation of a ground robot among movable obstacles,'' in \emph{2014 IEEE
  International Symposium on Safety, Security, and Rescue Robotics (2014)},
  2014, pp. 1--8.

\bibitem{collins2021scalable}
L.~Collins, P.~Ghassemi, E.~T. Esfahani, D.~Doermann, K.~Dantu, and
  S.~Chowdhury, ``Scalable coverage path planning of multi-robot teams for
  monitoring non-convex areas,'' in \emph{2021 IEEE International Conference on
  Robotics and Automation (ICRA)}.\hskip 1em plus 0.5em minus 0.4em\relax IEEE,
  2021, pp. 7393--7399.

\bibitem{ropero2019terra}
F.~Ropero, P.~Mu{\~n}oz, and M.~D. R-Moreno, ``Terra: A path planning algorithm
  for cooperative ugv--uav exploration,'' \emph{Engineering Applications of
  Artificial Intelligence}, vol.~78, pp. 260--272, 2019.

\bibitem{garciagarcia2017review}
A.~Garcia-Garcia, S.~Orts-Escolano, S.~Oprea, V.~Villena-Martinez, and
  J.~Garcia-Rodriguez, ``A review on deep learning techniques applied to
  semantic segmentation,'' \emph{arXiv preprint arXiv:1704.06857}, 2017.

\bibitem{ding2021review}
Y.~Ding, B.~Xin, and J.~Chen, ``A review of recent advances in coordination
  between unmanned aerial and ground vehicles,'' \emph{Unmanned Systems},
  vol.~9, no.~02, pp. 97--117, 2021.

\bibitem{qin2019autonomous}
H.~Qin \emph{et~al.}, ``Autonomous exploration and mapping system using
  heterogeneous uavs and ugvs in gps-denied environments,'' \emph{IEEE
  Transactions on Vehicular Technology}, vol.~68, no.~2, pp. 1339--1350, 2019.

\bibitem{delmerico2017active}
J.~Delmerico, E.~Mueggler, J.~Nitsch, and D.~Scaramuzza, ``Active autonomous
  aerial exploration for ground robot path planning,'' \emph{IEEE Robotics and
  Automation Letters}, vol.~2, no.~2, pp. 664--671, 2017.

\bibitem{su2021framework}
Z.~Su, C.~Wang, X.~Wu, Y.~Dong, J.~Ni, and B.~He, ``A framework of cooperative
  uav-ugv system for target tracking,'' in \emph{2021 IEEE International
  Conference on Real-time Computing and Robotics (RCAR)}.\hskip 1em plus 0.5em
  minus 0.4em\relax IEEE, 2021, pp. 1260--1265.

\bibitem{yue2020collaborative}
Y.~Yue \emph{et~al.}, ``Collaborative semantic understanding and mapping
  framework for autonomous systems,'' \emph{IEEE/ASME Transactions on
  Mechatronics}, vol.~26, no.~2, pp. 978--989, 2020.

\bibitem{grocholsky2006cooperative}
B.~Grocholsky, J.~Keller, V.~Kumar, and G.~Pappas, ``Cooperative air and ground
  surveillance,'' \emph{IEEE Robotics Automation Magazine}, vol.~13, no.~3, pp.
  16--25, 2006.

\bibitem{peterson2018online}
J.~Peterson, H.~Chaudhry, K.~Abdelatty, J.~Bird, and K.~Kochersberger, ``Online
  aerial terrain mapping for ground robot navigation,'' \emph{Sensors},
  vol.~18, no.~2, p. 630, 2018.

\bibitem{shah2022viking}
D.~Shah and S.~Levine, ``{ViKiNG: Vision-Based Kilometer-Scale Navigation with
  Geographic Hints},'' in \emph{Proceedings of Robotics: Science and Systems},
  2022.

\bibitem{alamouri2021development}
A.~Alamouri, M.~Hassan, and M.~Gerke, ``Development of a methodology for
  real-time retrieving and viewing of spatial data in emergency scenarios,''
  \emph{Applied Geomatics}, vol.~13, no.~4, pp. 747--761, 2021.

\bibitem{fedorenko2018global}
R.~Fedorenko, A.~Gabdullin, and A.~Fedorenko, ``Global ugv path planning on
  point cloud maps created by uav,'' in \emph{2018 3rd IEEE International
  Conference on Intelligent Transportation Engineering (ICITE)}, 2018, pp.
  253--258.

\bibitem{vandapel2006unmanned}
N.~Vandapel, R.~R. Donamukkala, and M.~Hebert, ``Unmanned ground vehicle
  navigation using aerial ladar data,'' \emph{The International Journal of
  Robotics Research}, vol.~25, no.~1, pp. 31--51, 2006.

\bibitem{kaslin2016collaborative}
R.~K{\"a}slin \emph{et~al.}, ``Collaborative localization of aerial and ground
  robots through elevation maps,'' in \emph{2016 IEEE International Symposium
  on Safety, Security, and Rescue Robotics (SSRR)}.\hskip 1em plus 0.5em minus
  0.4em\relax IEEE, 2016, pp. 284--290.

\bibitem{miller2021any}
I.~D. Miller \emph{et~al.}, ``Any way you look at it: Semantic crossview
  localization and mapping with lidar,'' \emph{IEEE Robotics and Automation
  Letters}, vol.~6, no.~2, pp. 2397--2404, 2021.

\bibitem{mox2020mobile}
D.~Mox \emph{et~al.}, ``Mobile wireless network infrastructure on demand,'' in
  \emph{2020 IEEE International Conference on Robotics and Automation
  (ICRA)}.\hskip 1em plus 0.5em minus 0.4em\relax IEEE, 2020, pp. 7726--7732.

\bibitem{nouri20213d}
N.~Nouri \emph{et~al.}, ``3d multi-uav placement and resource allocation for
  energy-efficient iot communication,'' \emph{IEEE Internet of Things Journal},
  2021.

\bibitem{cheng2018air}
N.~Cheng \emph{et~al.}, ``Air-ground integrated mobile edge networks:
  Architecture, challenges, and opportunities,'' \emph{IEEE Communications
  Magazine}, vol.~56, no.~8, pp. 26--32, 2018.

\bibitem{ginting2021chord}
M.~F. Ginting, K.~Otsu, J.~A. Edlund, J.~Gao, and A.-A. Agha-Mohammadi,
  ``Chord: Distributed data-sharing via hybrid ros 1 and 2 for multi-robot
  exploration of large-scale complex environments,'' \emph{IEEE Robotics and
  Automation Letters}, vol.~6, no.~3, pp. 5064--5071, 2021.

\bibitem{huang2019duckiefloat}
Y.-W. Huang \emph{et~al.}, ``Duckiefloat: a collision-tolerant
  resource-constrained blimp for long-term autonomy in subterranean
  environments,'' \emph{arXiv preprint arXiv:1910.14275}, 2019.

\bibitem{liu2022large}
X.~Liu \emph{et~al.}, ``Large-scale autonomous flight with real-time semantic
  slam under dense forest canopy,'' \emph{IEEE Robotics and Automation
  Letters}, vol.~7, no.~2, pp. 5512--5519, 2022.

\bibitem{quigley2019ovc}
M.~Quigley \emph{et~al.}, ``The open vision computer: An integrated sensing and
  compute system for mobile robots,'' in \emph{2019 International Conference on
  Robotics and Automation (ICRA)}.\hskip 1em plus 0.5em minus 0.4em\relax IEEE,
  2019, pp. 1834--1840.

\bibitem{hinzmann2017mapping}
T.~Hinzmann, J.~L. Sch{\"o}nberger, M.~Pollefeys, and R.~Siegwart, ``Mapping on
  the fly: Real-time 3d dense reconstruction, digital surface map and
  incremental orthomosaic generation for unmanned aerial vehicles,'' in
  \emph{Field and Service Robotics}.\hskip 1em plus 0.5em minus 0.4em\relax
  Springer, 2018, pp. 383--396.

\bibitem{kern2020openrealm}
A.~Kern, M.~Bobbe, Y.~Khedar, and U.~Bestmann, ``Openrealm: Real-time mapping
  for unmanned aerial vehicles,'' in \emph{2020 International Conference on
  Unmanned Aircraft Systems (ICUAS)}, 2020, pp. 902--911.

\bibitem{campos2021orb}
C.~Campos, R.~Elvira, J.~J.~G. Rodr{\'\i}guez, J.~M. Montiel, and J.~D.
  Tard{\'o}s, ``Orb-slam3: An accurate open-source library for visual,
  visual--inertial, and multimap slam,'' \emph{IEEE Transactions on Robotics},
  vol.~37, no.~6, pp. 1874--1890, 2021.

\bibitem{dellaert2012factor}
F.~Dellaert, ``Factor graphs and gtsam: A hands-on introduction,'' Georgia
  Institute of Technology, Tech. Rep., 2012.

\bibitem{romera2018erfnet}
E.~Romera, J.~M. Álvarez, L.~M. Bergasa, and R.~Arroyo, ``Erfnet: Efficient
  residual factorized convnet for real-time semantic segmentation,'' \emph{IEEE
  Transactions on Intelligent Transportation Systems}, vol.~19, no.~1, pp.
  263--272, 2018.

\bibitem{wang2020deep}
J.~Wang \emph{et~al.}, ``Deep high-resolution representation learning for
  visual recognition,'' \emph{IEEE transactions on pattern analysis and machine
  intelligence}, vol.~43, no.~10, pp. 3349--3364, 2020.

\bibitem{cowley2021upslam}
A.~Cowley, I.~D. Miller, and C.~J. Taylor, ``Upslam: Union of panoramas slam,''
  in \emph{2021 IEEE International Conference on Robotics and Automation
  (ICRA)}, 2021, pp. 1103--1109.

\bibitem{RFC6126}
J.~Chroboczek, ``The babel routing protocol,'' Internet Requests for Comments,
  RFC Editor, RFC 6126, April 2011.

\bibitem{miller2020mine}
I.~D. Miller \emph{et~al.}, ``Mine tunnel exploration using multiple
  quadrupedal robots,'' \emph{IEEE Robotics and Automation Letters}, vol.~5,
  no.~2, pp. 2840--2847, 2020.

\bibitem{redmon2018yolov3}
J.~Redmon and A.~Farhadi, ``Yolov3: An incremental improvement,'' \emph{arXiv
  preprint arXiv:1804.02767}, 2018.

\end{thebibliography}

\end{document}